%% file: acl.tex
\pdfoutput=1

\documentclass[11pt]{article}

\usepackage[final]{acl}

\usepackage{times}
\usepackage{latexsym}

\usepackage[T1]{fontenc}

\usepackage[utf8]{inputenc}

\usepackage{microtype}

\usepackage{inconsolata}

\usepackage{graphicx}

\usepackage{times}
\usepackage{latexsym}

\usepackage[T1]{fontenc}

\usepackage[utf8]{inputenc}

\usepackage{microtype}

\usepackage{algorithm}
\usepackage{amsmath}
\usepackage{algorithmic}

\title{Interactively Learning
Social Media Representations Improves News Source Factuality Detection}

\usepackage{lipsum}

\author{Nikhil Mehta \\ Department of Computer Science \\ Purdue University \\ West Lafayette, IN 47907 \\ {\tt mehta52@purdue.edu} \\\And Dan Goldwasser \\ Department of Computer Science \\ Purdue University \\ West Lafayette, IN 47907 \\ {\tt dgoldwas@purdue.edu} \\}

\begin{document}
\maketitle
\begin{abstract}
The rise of social media has enabled the widespread propagation of fake news, text that is published with an intent to spread misinformation and sway beliefs. Rapidly detecting fake news, especially as new events arise, is important to prevent misinformation.

While prior works have tackled this problem using supervised learning systems, automatedly modeling the complexities of the social media landscape that enables the spread of fake news is challenging. On the contrary, having humans fact check all news is not scalable. Thus, in this paper, we propose to approach this problem \textit{interactively}, where humans can interact to help an automated system learn a better social media representation quality. On real world events, our experiments show performance improvements in detecting factuality of news sources, even after few human interactions.

\end{abstract}

\section{Introduction}
\label{sec:intro}
\input{introduction.tex}

 \section{Related Work}
 \label{sec:related_work}
 \input{related.tex}

 \section{Graph Model}
 \label{sec:model}
 \input{model.tex}

 \section{Interactive Protocol}
 \label{sec:interactions}
 \input{interactions.tex}

 \section{Experiments}
 \label{sec:experiments}
 \input{experiments.tex}
 
 \section{Discussion}
 \label{sec:discussion}
 \input{discussion.tex}
 
 \section{Summary and Future Work}
 \label{sec:summary}
 \input{summary.tex}

 \section{Acknowledgements}
 \label{sec:acknowledgements}
 \input{acknowledgements.tex}
 
 \section{Ethics Statement}
 In this section, we first discuss some limitations of our model (\ref{sec:limitations}), and then expand on that with a discussion on ethics as it relates to our data collection, data usage, human interaction, and the deployment of our models (\ref{sec:ethics}).
 
 \subsection{Limitations}
 \label{sec:limitations}
 This work tackles fake news source detection in English on Twitter (our social media platform of choice). Our methods may or may not apply to other languages with different morphology, and other social networking platforms. We leave the investigation of this to future work, but are optimistic that especially with the benefit of interactivity, our methods may generalize.
 
 The nature of our interactive framework also requires human interactors to interact, which could be a potential limitation. Interactors must have some general understanding of news content and be able to identify if two entities (users, sources, or articles) have similar content relationships. However, as interactors are just looking for content/perspective similarity, they need not be aware of the latest events or be fake news detection specialists. Further, human interactors don't analyze user-specific information or profile users themselves, they just determine if users have similar content relationships. 
 
 We used a single GeForce GTX 1080 NVIDIA GPU to train our models, with 12 GB of memory. As our models are largely textual based, they do not require much GPU usage. However, scaling our experiments to larger scale settings in real world settings could require more compute, which may be a potential limitation. Our hyper-parameter search, mentioned in App~\ref{sec:experimental_settings} was done manually.
 
 \subsection{Ethics}
 \label{sec:ethics}
 To the best of our knowledge no code of ethics was violated throughout the experiments done in this paper. We reported all hyper-parameters and other technical details necessary to reproduce our results, and release the code and dataset we collected. We evaluated our model on two  datasets that we collected in this paper, and was collected by prior work, but it is possible that results may differ on other datasets. We believe our methodology is solid and applies to any social media fake news setting. Due to lack of space, we placed some of the technical details and discussion to the Appendix section. The results we reported supports our claims in this paper and we believe it is reproducible. Any qualitative result we report is an outcome from a machine learning model that does not represent the authors' personal views. For anything associated with the data we use, we do not include account information and all results are anonymous.
 
 In our dataset release, we include sources, users, and articles, with enough data to produce the results described in the paper and the Appendix. Sources are public information provided in \cite{baly:2020:ACL2020}, and we map each to an ID. We release article graph embeddings, which can be used to train our models. As these embeddings are neural network representations, they can't be mapped back to article text. However, we also release article URLs, so that the articles can be downloaded, if they are still publicly available. Additionally, we release the Twitter data that we used, in complicance with the Twitter API policies \footnote{\url{https://developer.twitter.com/en/developer-terms/more-on-restricted-use-cases}}. In our dataset release, each user is referenced by their Twitter ID, and their graph ID (the graph ID is meaningless on it's own). We release the mapping of the Twitter ID to the graph ID.  By us only releasing Twitter ID's, and not the actual Twitter text or user information, in order to download the exact Twitter data that we used, users must use the Twitter API to gather the latest public information\footnote{\url{https://developer.twitter.com/en/docs/twitter-api}}. This ensures that we respect user privacy, in accordance with the policies mentioned by the Twitter API, as only user content that is still public can be downloaded and we are not storing/releasing any data. We also provide the model representations for each user, article, and source we used as our initial embedding in the graph. As these are neural network model embeddings, they can't be mapped back to the individual text. Our data is meant for academic research purposes and should not be used outside of academic research contexts. All our data is in English.

 In this paper, we did not use any of the Twitter data for user surveillance purposes, and we encourage the community to do the same, to respect user privacy. We also do not profile users, we only use the user insights as an aggregate to classify news sources. Further, we only use public Twitter profiles, which there are enough of for our framework to work in real-time situations. When doing human interactions, we show humans public Twitter information, so that they can determine user similarity. To do this, we use the Twitter API to determine the Twitter data that is publicly available at the time of interaction, show that to humans, and then discard the Twitter information. Further, in our graph model, we do not store any user-specific information, we only store neural network model embeddings which are used for training and cannot be mapped back to the original text or user. The same is true for articles, so we we are actually discarding all the text (Twitter, article, and source). Users of our framework should also do the same - use public knowledge for interactions and not store any user/article specific data, rather use the appropriate APIs to retrieve the data when needed. 
 
 Our framework in general is intended to be used to defend against fake news. While our framework could be used to build better methods of designing fake news, our methodology of interactive fake news detection could guard against that as well. We caution that our models and methods be considered and used carefully. This is because in an area like fake news detection, there are great consequences of wrong model decisions, such as unfair censorship and other social related issues. Further, despite our efforts, it is possible our models are biased, and this should also be taken into consideration. Our protocol of building sub-graphs based on model confusion that we used when showing humans what to interact on, can be used to get insights into the model to help prevent some of these issues as well. However, this is definitely an area of future work.
 
 In the interactive setting we proposed, our approach relies on getting insights from human interactors and using that to improve performance in fake news detection. While that lead to performance improvements in this work and we believe it will hold in different settings, there could be issues, such as biased humans. Running interactions at large scale with multiple human experts per sub-graph can help mitigate some of these issues. For example, edges can be weighted in the graph based on how many humans chose to add them. Thus, extremely biased interactors decisions would be given less weight, and maybe even not considered by the model. We leave this for future work. However, despite this, there may still be some human interactor bias that can leak into the final fake news detection model, which is why perhaps important decisions should not be made only by machine learning models, but rather the models be used as a tool. 

 As mentioned in the Appendix~\ref{sec:additional_interactor_details}, the human interactors we used were Compute Science PhD students. The interactors were awarded research credits for their work, as the hours they spent working on the task were considered as part of their research credit hours. They were explained the entire process before hand including what the interactions would be used for, and agreed to perform the interaction. The total interaction process took under 3 hours, including the time spent explaining the process.
 
 These and many other issues are things to consider when using fake news detection models such as the one proposed in this work.

\bibliography{acl}
\bibliographystyle{acl_natbib}

\newpage
\appendix

\section{Supplemental Material: Fake News Source Detection}
\label{sec:supplemental}
In this section, we provide implementation details for our models for fake news source detection. The original dataset we used has 859 sources: 452 \textit{high} factuality, 245 \textit{mixed}, and 162 \textit{low}, and was released publicly by \cite{baly:2020:ACL2020}\footnote{https://github.com/ramybaly/News-Media-Reliability}. We then extended it by scraping sources from the Media Bias/Fact Check website\footnote{https://mediabiasfactcheck.com}, to gain better coverage of more recent events. The dataset does not include any other raw data (articles, sources, etc.), so we must scrape our own.

\subsection{Data Collection}
\label{appendix_data_collection}
We will release the data and code for this paper upon acceptance. Our data collection process is identical to \citeauthor{mehta2022tackling} as we use their code, but we briefly describe the process here completion. Further details are available in \citeauthor{mehta2022tackling}

Following the process used in \citeauthor{mehta2022tackling}, we tried to scrape news articles for each source in the dataset using public libraries (Newspaper3K \footnote{https://github.com/codelucas/newspaper}, news-please \footnote{https://github.com/fhamborg/news-please} \cite{Hamborg2017}, and Scrapy \footnote{https://github.com/scrapy/scrapy}). In cases where the webpage of the news source was removed as often happens with fake news websites, we used the Wayback Machine \footnote{https://archive.org/web/} to download the articles, if possible. As explained in Sec~\ref{sec:model}, we attempted to get up to 300 articles for each source. For the sources we got for the \cite{baly:2020:ACL2020} dataset, our statistics are the same as \citeauthor{mehta2022tackling} as we use their data, and thus the sources have an average of 109 articles with a STD of 36.

To scrape Twitter data, we used the Twitter API\footnote{https://developer.twitter.com/en/docs}. In order to densely populate the graph, we attempted to scrape up to 5000 followers for each source that had a Twitter account ((72.5\% of the sources, the same number as \cite{baly:2020:ACL2020, mehta2022tackling}). Further, to find users that propagate articles, we used the Twitter Search API to search articles. From the returned Twitter results, we use users that mention the article title or the article URL and post their tweet within 3 months of the original article being published. For each user found, we download their profile and add them to our graph, making the appropriate Twitter User to Article connection as discussed in Sec~\ref{sec:model}. Finally, we scraped the followers of each Twitter user in the graph, and connected them to any user they followed that was in the graph. This increases the connectivity of the graph and allows us to better capture the social media landscape. To maintain a densely connected graph, we remove users that do not connect to any other node in the graph.

To get the data for the YouTube embeddings for the sources, we used the ones released publicly by \cite{baly:2020:ACL2020}, who were able to scrape YouTube channels for 49\% of sources.

\subsection{Event Collection}
\label{sec:event_collection}
To collect the data for each event ($E1$ from June 2, 2019 - Jan 1, 2021 and $E2$ from Feb 2, 2021 - May 6, 2022), we filtered by date and downloaded Tweets mentioning certain hashtags and a URL with one of the sources in the dataset. The hashtags/search terms we used to collect data for the \textit{Black Lives Matter} event were: \textit{Black Lives Matter, BLM, blacklivesmatter, Floyd, George Floyd}. 

We then filtered the data to find the top \textit{high}, \textit{mixed}, and \textit{low} factuality sources that were mentioned on Twitter for each of these events and time periods. We kept sources that had at least 10 articles with Twitter users propagating them. We ended up with at least 33 sources for each factuality level in each event data split. Other sources that were in the training set of \citeauthor{baly:2020:ACL2020} were used to train the initial Graph Embedding. Detailed statistics for the training set and the sources used in one of our Black Lives Matter splits are shown in Table~\ref{table:data_statistics}.

\subsection{Experimental Settings}
\label{sec:experimental_settings}
\subsubsection{Initial Embeddings}
\label{subsec:initial_embeddings}
We used the same initial graph node embedding representations as \citeauthor{mehta2022tackling}, which we briefly explain here. The Twitter embedding we used for each source and each Twitter user was a 773 dimensional vector consisting of the SBERT \cite{reimers2019sentence} (RoBERTa \cite{liu2019roberta} Base NLI model) representation of their bio (up to the first 512 tokens) concatenated with a variety of numerical features, as follows: (1) a binary number representing whether the source is verified, (2) the number of users a source follows and the number that follow it, (3) the number of tweets the user posts, and (4) the number of favorites/likes the users' tweets have received. The YouTube embedding we used consisted of the following numerical features: the average of the number of views, the number of dislikes, and the number of comments for each video the source posted on YouTube. For articles, we used the same SBERT RoBERTa model to generate an embedding based on the article text, which ends up being a 768 dimensional vector. In all cases where we encoded text for embedding representations, we use SBERT \cite{reimers2019sentence} RoBERTa \cite{liu2019roberta} model because it provides semantically meaningful sentence representations for the text. 

\subsubsection{Models and Training}

We used the publicly released code of \citeauthor{mehta2022tackling}, which was built using PyTorch \cite{NEURIPS2019_9015} and DGL (Deep Graph Library) \cite{wang2019deep} in Python. The R-GCN used consists of 5 layers, 128 hidden units, a learning rate of 0.001, and a batch size of 128 for Node Classification. The initial source and article embeddings have hidden dimension 768, while the user one has dimension 773. To do 3-way source classification, the final fully connected layer has size 3.

We trained our models using a 12GB TITAN XP GPU card. To learn the initial model which was used to determine where to perform interactions, it took approximately 2 hours. Training after interactions takes approximately 30 minutes total. Inductive settings do not have any training, and take minutes to run as we only have to compute embeddings for nodes that we are attempting to classify.

We used the development set to evaluate model performance, and choose the best hyper-parameters.

\section{Supplemental Material: Interaction Graphs}
\label{appendix:supplemental_advice_graphs}
In this section, we provide details about the interaction graphs we showed users in \ref{appendix:interaction_graph_details}. Then, in \ref{appendix_human_interaction_details}, we discuss the interaction process (\ref{appendix:interaction_process}), the details behind the interactions used in the main paper (\ref{sec:initial_interactor_results}), and finally new interactions  on the second Black Lives Matter Split (\ref{sec:additional_interactor_details}).

\begin{table*}
\begin{center}
\begin{tabular}{|p{6.25cm}|p{0.9cm}|p{0.9cm}|p{0.9cm}|p{0.9cm}|p{0.9cm}|p{0.9cm}|p{0.9cm}|}
  \hline
  {\textbf{\small Model}} & {\textbf{\small T1-1 Acc}}
  & {\textbf{\small T1-1 F1}} & {\textbf{\small T1-2 Acc}} & {\textbf{\small T1-2 F1}} & {\textbf{\small T2-1 Acc}} &  {\textbf{\small T2-1 F1}} & {\textbf{\small \# Edges}} \\

 \hline
  \small No Interactions & \small 41.79  & \small 37.10  & \small 41.93 & \small 35.95  & \small 37.50 & \small 33.54 & \small - \\
  \small No Interactions Train & \small 85.71  & \small 85.16 &  \small 68.42 & \small 63.00 & \small 40.40 & \small 30.75 & \small - \\
  \hline 
  \small Protocol 1: Human Interactions in T1-1 & \small 43.28   & \small  37.39 & \small 45.16 & \small 42.34 & \small - & \small - & \small 65 \\
  \small Protocol 2: Human Interactions in T1-1 & \small 71.68 & \small 72.50 & \small52.63 & \small 41.82 & \small 41.41 & \small 34.79 & \small 65 \\

 \hline
\end{tabular}
\end{center}
\captionsetup{justification=centering}
\caption{\small Additional interaction results across the two of our protocols. With minimal interactions, we still see performance improvements in the fully inductive setting in Protocol 1 using all the interactions done by the three humans. We also see how the interactions allow us to learn a better model in Protocol 2, as seen by the improvements on $T2$-$1$, despite no interactions performed there (interactions were only on $T1$-$1$).}
\label{table:additional_interactions}
\end{table*}

\begin{table}[t]
\begin{center}
\begin{tabular}{|p{2.0cm}|p{1cm}|p{1cm}|p{1cm}|}
  \hline
  {\textbf{\small Dataset}} & {\textbf{\small Low}} & {\textbf{\small Mixed}} & {\textbf{\small High}} \\

 \hline
  \small Training Event & \small 57 &  \small 81  &   \small 153 \\
  \hline
  \small T1 & \small 56 &  \small 33 &  \small 33 \\
  \hline 
  \small T2  & \small  33 &  \small 33 &  \small 33 \\
 \hline
\end{tabular}
\end{center}
\vspace{-10pt}
\captionsetup{justification=centering}
\caption{Number of sources in our datasets for the training event (added to the training set to train the initial model), T1 (first event) and T2 (second event). The results on these data sets are shown in Table~\ref{table:additional_interactions}.}
\vspace{-10pt}
\label{table:data_statistics}
\end{table}

\subsection{Interaction Graph Details}
\label{appendix:interaction_graph_details}
\begin{figure*}
    \centering
    \includegraphics[width=1.0\textwidth]{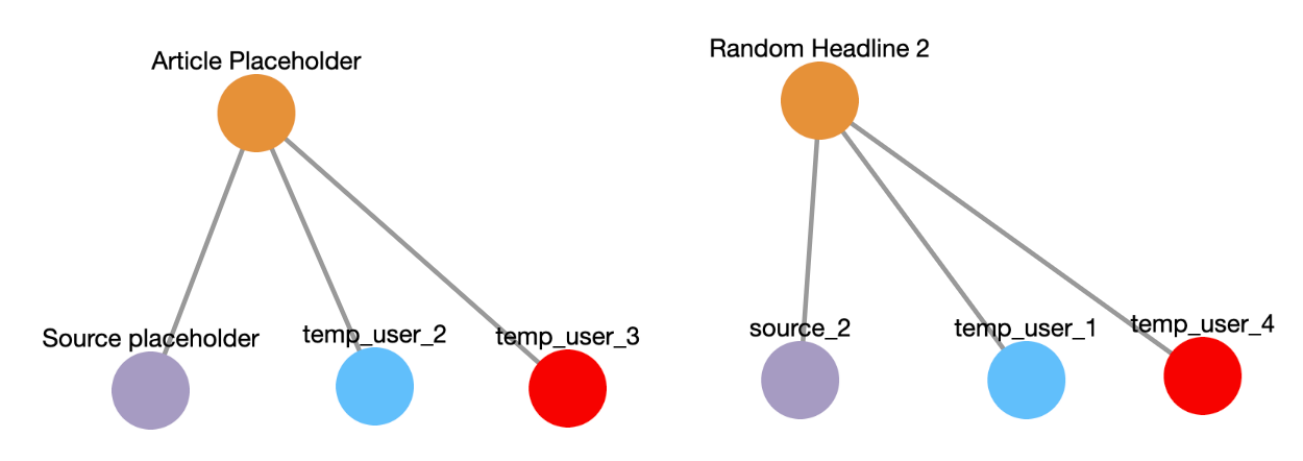}
    \caption{Example of an interaction graph that has been anonymized. The two red nodes are the pairs of users that were identified in Sec~\ref{sec:getting_human_interactions}, shown by the Twitter usernames. The two orange nodes are the article nodes, and shown by their headlines without determiners. Blue nodes are other users that propagate the same articles (could be celebrities - users with over 1000 followers, and purple nodes are sources)}. 
    \label{fig:interaction_graph_example}
    \vspace{-8pt}
\end{figure*}

The sub-graphs were constructed by first picking pairs of users, and then adding context around them, as discussed in Sec~\ref{sec:getting_human_interactions}. The context consists of the article each user propagated, the sources that published it, other users that propagated the same articles, and finally some additional celebrities (users with more than 1000 followers) that are followed by one of the users in the graph. All articles added to the graph have to be about the same event, and they are found by searching hashtags related to the event (in our case Black Lives Matter or Climate Change, hashtags below) in a four week span on Twitter. An example of an interaction graph that is shown to humans is in Fig.~\ref{fig:interaction_graph_example}

\begin{figure*}
    \centering
    \includegraphics[width=1.0\textwidth]{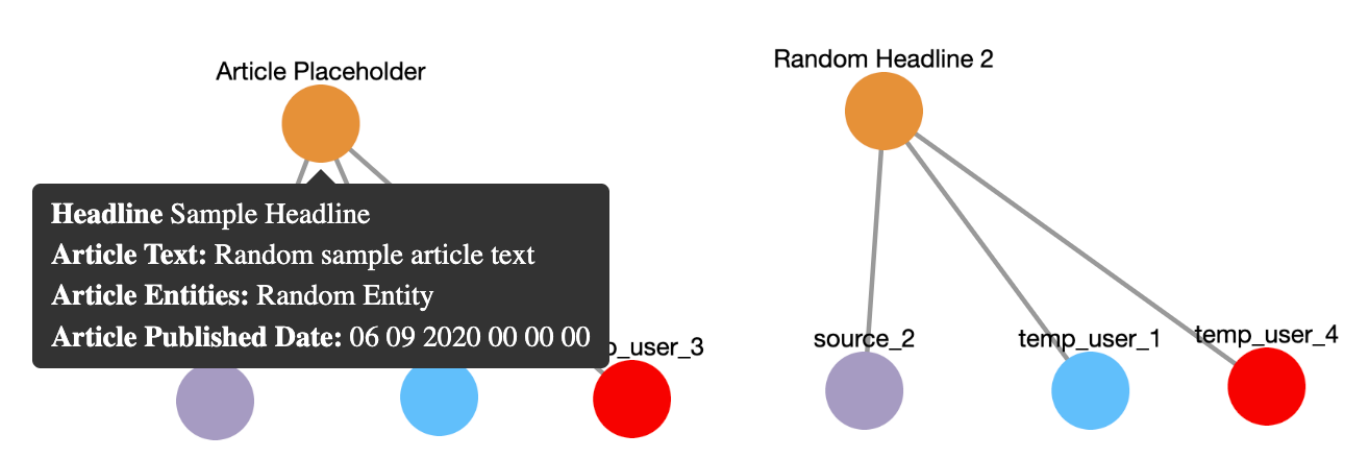}
    \caption{Example of an interaction graph where we can see metadata about an article, by clicking on the article node. This would be filled in during real human interactions, to allow humans to analyze the article and the context around it, but is currently anonymized.}. 
    \label{fig:interaction_graph_example_article}
    \vspace{-8pt}
\end{figure*}

\begin{figure*}
    \centering
    \includegraphics[width=1.0\textwidth]{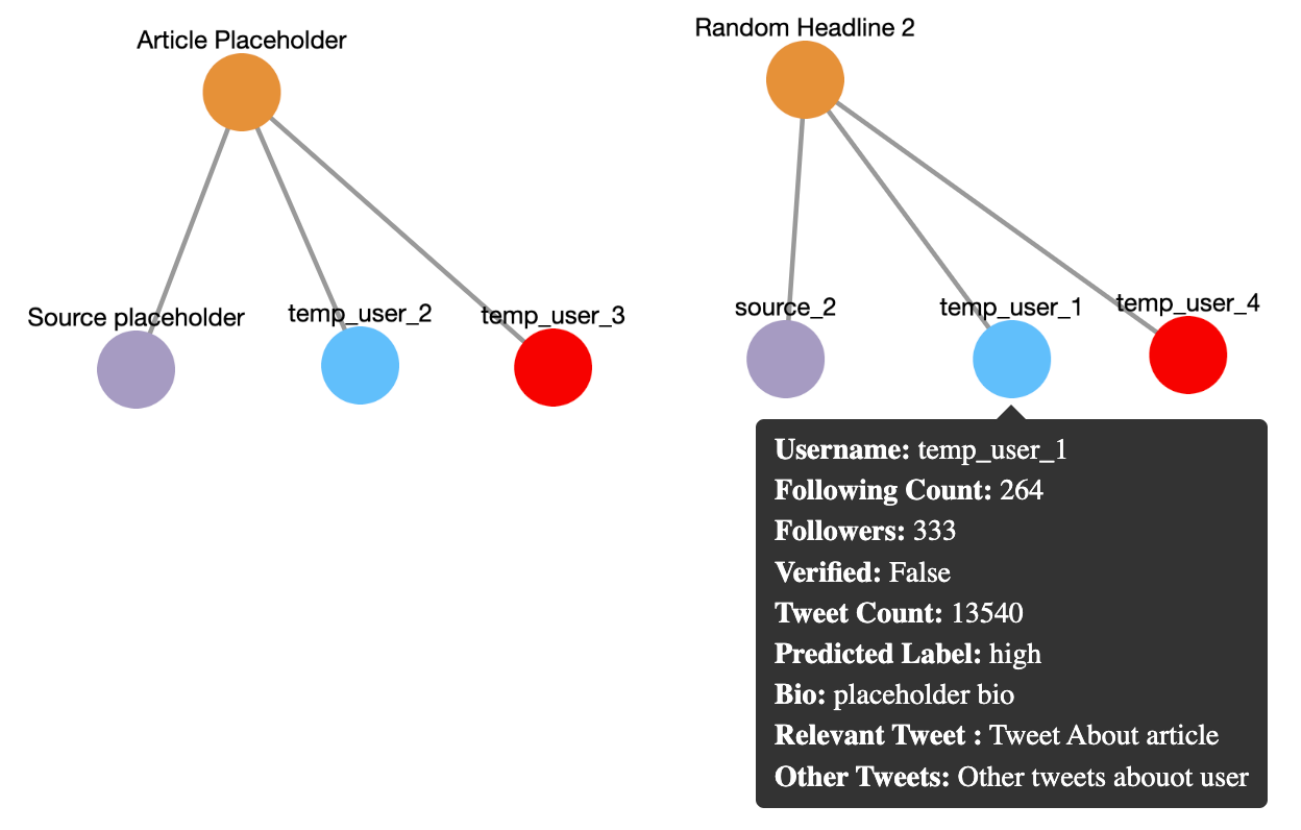}
    \caption{Example of an interaction graph where we can see metadata about a user, by clicking on the user node. This would be filled in (with data from Twitter) during real human interactions, to allow humans to analyze the user and the context around it, but is currently anonymized. Source nodes with Twitter profiles would appear with the same metadata.}. 
    \label{fig:interaction_graph_example_user}
    \vspace{-8pt}
\end{figure*}

Each node in the interaction graphs also contains metadata to provide more information to the human interactors. As seen in Fig~\ref{fig:interaction_graph_example_article}, article nodes consist of the headline, article text, article entities, and the date the article was published. As seen in Fig~\ref{fig:interaction_graph_example_user} source and user nodes contain Twitter information such as: username, following count, number of followers, whether they are verified, how many tweets they make, what is their model predicted label, their biography, the tweet they made about the article, and other tweets they made about the same event.  

Once the graphs are built, we show them to humans and ask a series of questions to guide the interaction process. Each question asks the user to create an edge in the interaction graph based if there is a positive relationship between the two nodes. A Positive relationships mean the nodes have similar content preferences. If a positive relationship doesn't exist, or there is not enough data to clearly determine a positive relationship, humans are asked to not make any edge connections. Thus, humans are asked to ignore potentially subjective cases. Humans identifying positive relationships has multiple benefits for fake news detection: \textbf{(1) Simplicity:} It is far simpler than identifying factuality, so it can be used to detect fake news quickly. The simplicity of the interaction process is due to the fact that interactors only have to read a small amount of content (a few user tweets/profile information + up to two article headlines/summaries), compared to reading multiple articles and gaining real world knowledge. On average, human interactors spent 3 minutes per interaction graph, and made an average of 8 connections in this time. \textbf{(2) Effectiveness:} Interactions improve social media representation quality and thus social homophily, and that's what leads to performance improvements.  Here are the questions we asked human interactors for each sub-graph shown:

\begin{enumerate}
    \item Are there any users that are similar to each other? Please connect them.
    \item Are there any articles that are similar to each other? Please connect them.
    \item Are any users likely to propagate any of the articles? Please connect them to the appropriate article.
    \item Are any users likely to interact with another user? Please connect those pairs of users.
    \item Are any users likely to interact with any sources? Please connect those users of the respective source.
\end{enumerate}

 \subsection{Human Interactor Details}
 \label{appendix_human_interaction_details}
 In this sub-section, we first discuss the interaction process, including the graphical interface we built for this task (\ref{appendix:interaction_process}). Then, we discuss the two rounds of human interaction protocols we did for Black Lives Matter. Climate change details follow in Sec~\ref{sec:climate_change_results}. The first split of Black Lives Matter, Sec~\ref{sec:initial_interactor_results} was presented in the main paper, and the second split \ref{sec:additional_interactor_details} appears in this section.
 
 \subsubsection{Interaction Process}
 \label{appendix:interaction_process}
 The human interactors use a graphical interface that displays the interaction graphs. We hosted the interface on a website built for this interaction process. Humans must answer the questions discussed above (Sec~\ref{appendix:interaction_graph_details}) by connecting nodes to create edges, which are then saved on our server and can be incorporated into the broad event graph as new edges, when evaluating the overall performance. The examples of the graphs human interactors see when interacting, and the metadata they are provided with, can be seen in Fig~\ref{fig:interaction_graph_example}, Fig~\ref{fig:interaction_graph_example_article}, and Fig~\ref{fig:interaction_graph_example_user}. Examples of connections made are in ~\ref{sec:human_annotation_analysis}. 
 
 \subsubsection{Initial Interactor Details}
 \label{sec:initial_interactor_results}
 We now describe the initial interaction session we discussed in the main Paper on Black Lives Matter and used for all the BLM Human results presented there. The interactor used for this session was an Asian-American PhD student in Computer Science and Natural Language Processing. They were explained of the entire process before hand including what the interactions would be used for, and agreed to perform the interaction. They interacted on 20 graphs per data split (E1-1, dev, and E2-1), which took under an hour for each split.

\begin{table*}
\begin{center}
\begin{tabular}{|p{5.2cm}|p{0.7cm}|p{0.7cm}|p{0.7cm}|p{0.7cm}|p{0.7cm}|p{0.7cm}|p{0.7cm}|p{0.7cm}|p{0.7cm}|}
  \hline
  {\textbf{\small Model}} & {\textbf{\small E1-1 Acc}}
  & {\textbf{\small E1-1 F1}} & {\textbf{\small E1-2 Acc}} & {\textbf{\small E1-2 F1}} & {\textbf{\small E2-1 Acc}} &
  {\textbf{\small E2-1 F1}} &
  {\textbf{\small E2-2 Acc}} &
  {\textbf{\small E2-2 F1}} &{\textbf{\small \# Edges}} \\

 \hline
  \small CLM No Interactions & \small 40.16  & \small 32.77 & \small 39.65 & \small 31.86 & \small 34.88 & \small 30.93 & \small - & \small - & \small - \\
  \small CLM Sim. Interactions on 100\% of Data in E1-1 + E2-1 & \small 46.72 & \small 41.59 & \small 39.65 & \small 31.86 & \small 44.18 & \small 41.57 & \small - & \small - & \small 12,602 \\
  \small CLM Human Interactions in E1-1 + E2-1 & \small 46.72 & \small 43.94 & \small 39.65 & \small 31.86 & \small 39.53 & \small 36.95 & \small - & \small - & \small 47 \\
  \hline 
  \small CLM No Interactions Train & \small - & \small - & \small 49.29  & \small 44.84 & \small 44.77 & \small 42.35 & \small - & \small - & \small - \\
  \small CLM Sim. Interactions on 100\% of Data in E1-1 & \small - & - & \small 56.33 & \small 50.02  & \small 44.18 & \small 43.10 & \small - & \small -  & \small 12,602 \\
  \small CLM Human Interactions in E1-1 & \small  - & \small - & \small 53.52 &  \small 44.53 & \small 40.29 & \small 46.38 & \small - & \small - & \small 47 \\
  \hline 
  \small CLM No Interactions Train & \small - & \small - & \small 49.29  & \small 44.84 & \small 44.77 & \small 42.35 & \small 44.44 & \small 33.06 & \small - \\
  \small CLM Sim. Interactions on 100\% of Data in E1-1 + E1-2 & \small - & \small - & \small 56.33 & \small 50.02 & \small 41.86 & \small 36.42 & \small 34.04 & \small 24.93 & \small 12,602 \\
  \small CLM Human Interactions in E1-1 + E1-2 & \small - & \small - & \small 53.52  & \small 44.53 & \small 53.48 & \small 43.07 & \small 46.80 & \small 38.73 & \small 47 \\

 \hline
\end{tabular}
\end{center}
\vspace{-5pt}
\captionsetup{justification=centering}
\caption{\small Climate Change Results: Key: E1 and E2 are the two inductive graphs. E1-1/E2-1 is the first half, and E1-2/E2-2 are the second half. \# Edges shows the number of edges added by interactions to E1-1. \textbf{Protocol 1:} The top third refers to Protocol 1, where interactions result in performance improvements in the difficult, inductive, no training setting. In particular, we see improvements of 6.56\% Acc. and 11.17\% F1 on $E1$-$1$. We do not evaluate on $E2$-$2$, as no interactions are done on that split, so performance does not change. \textbf{Protocol 2:} The middle third refers to Protocol 2, where interactions result in performance improvements when we train on the interactions, and then apply the model to a new event with no interactions done. We do not evaluate on $E1$-$1$, as it is the training set), and $E2$-$2$, as no interactions are done and performance does not change. \textbf{Protocol 3:} The last third refers to Protocol 3, where interactions result in performance improvements when we train on the interactions and then do more interactions in the inductive setting ($E2$-$1$). We also see improvements in $E2$-$2$. We don't evaluate on $E1$-$1$ as it is the training set.}
\label{table:clm_results}
\end{table*}
 
 \subsubsection{Additional Interactor Details}
 \label{sec:additional_interactor_details}
  To expand our human interactor sessions, we also ran additional interaction sessions on the Black Lives Matter dataset (and Climate Change in Sec~\ref{sec:climate_change_results}. For variety, we used a different source split in this setting, that we will also release. For this reason, these additional results are not comparable to the ones in the main paper, and we refer to the new events as $T1$, $T1$-$1$, $T2$, etc. instead of $E1$, $E1$-$1$, $E2$, etc. The data collection was the same as before.
 
  For these additional interactions, we used three interactors of Asian descent, all fluent in the English language and all Computer Science PhD students. The interactors were awarded research credits for their work, as the hours that they spent working on the task were considered as part of their research credit hours. They were explained of the entire process before hand including what the interactions would be used for, and agreed to perform the interaction. Each interactor interacted on 10 graphs for $T1$-$1$, and spent less than one hour. A majority of the time was spent becoming familiar the interaction process, and once complete the interactions went more rapidly. As a test, we had one interactor do interactions on 10 more graphs, and they were able to do 10 graphs in less than 30 minutes, showing how this process can be done rapidly. Moreover, interactors spent an average of 3 mins. per interaction sub-graph, once familiar with the process. Across the three interactors, we had 65 unique edge connections made for $T1$-$1$. We had 31 edges that were repeated across the interactors, showing a reasonable level of interactor agreement given the the task. 
  
  Results for this additional interaction process is in Tab.~\ref{table:additional_interactions}, and they are consistent with the results for the Protocols in the main paper for the single interactor. For Protocol 1, we see improvements in the inductive setting on both the interaction and non-interaction half of $T1$. Thus, interactions help performance even when there is no additional training. Protocol 2 also leads to improvements and shows how interactions (done on $T1$-$1$) allow us to learn a better model for when no interactions are done (dev. set performance was 43.45\% Accuracy). This further shows how interactions can help to build a stronger model, especially on emerging news events. In addition, it is likely that more interactions would lead to more improvements, and we leave this for future work.

\section{Supplemental Material: Climate Change}
\label{sec:climate_change_results}
In this section, we expand upon the Climate Change (CLM) dataset results discussed first in Sec~\ref{sec:experiments}. We first discuss what search terms we used to collect the data for Climate Change in Sec~\ref{sec:clm_data_collect}. Then, in Sec~\ref{sec:clm_human_inter}, we explain the interaction process that was used for Climate Change, and the agreement statistics associated with it. Finally, in Sec~\ref{sec:clm_results}, we present detailed results for Climate Change, including simulated interactions. All results and conclusions are comparable with the Black Lives Matter results, showing that our approach generalizes across events.

\subsection{Data Collection}
\label{sec:clm_data_collect}

\begin{table}[t]
\begin{center}
\begin{tabular}{|p{2.0cm}|p{1cm}|p{1cm}|p{1cm}|}
  \hline
  {\textbf{\small Dataset}} & {\textbf{\small Low}} & {\textbf{\small Mixed}} & {\textbf{\small High}} \\
  \hline
  \small E1-1 & \small 30 &  \small 43 &  \small 49 \\
  \small E1-2 & \small 28 & \small 41 &  \small  47 \\
  \hline 
  \small E2-1 & \small 11 &  \small 15  &
  \small 17 \\
  \small E2-2 & \small 13 &  \small 16 &
  \small 18 \\
 \hline
\end{tabular}
\end{center}
\vspace{-10pt}
\captionsetup{justification=centering}
\caption{Number of sources in our dataset for climate change.}
\vspace{-10pt}
\label{table:data_statistics_climate}
\end{table}

We used the same Data Collection process for Climate Change as Black Lives Matter, discussed in Sec~\ref{sec:dataset} and Sec~\ref{appendix_data_collection}. This means we have the same 3 time periods for Climate Change as Black Lives Matter (BLM). The only difference between BLM and Climate Change are the search terms we used to search Twitter to collect the data. For Climate Change, we used the following search terms: \textit{fracking}, \textit{global warming}, \textit{climate change}, \textit{\#savetheplanet}, \textit{\#savethetrees}, \textit{\#climatechangeisreal},  \textit{\#waterpollution}, and \textit{\#climatestrike}. Statistics for the number of sources in each data split and their high, mixed, and low factuality distribution are shown in Table~\ref{table:data_statistics_climate}.

\subsection{Human Interaction Process}
\label{sec:clm_human_inter}
For the human interactions done on Climate Change, we used two male human interactors. Both are Computer Science P.h.D. students in Natural Language Processing of Asian descent. The interactions were done on 8 sub-graphs for each time period ($E1$-$1$, $E2$-$1$, and the development set), for a total of 24 sub-graphs interacted on. As with Black Lives Matter, interactors spent an average of about 3 minutes on each interaction sub-graph.

\subsection{Climate Change Results}
\label{sec:clm_results}

In-depth results for climate change are presented in Table~\ref{table:clm_results}. In the main paper, due to lack of space, we presented only baseline and human interaction results, which we expand upon here also showing simulated results. We can see that results improve and are consistent with Black Lives Matter Results for Protocols 1, 2, and 3. Protocol 1 (top third of the table) shows significant Accuracy and F1 improvements in the fully inductive setting, showing the power of minimal human interactions in the right places to improve the model without any training. Protocol 2 (middle third of the table) shows how interactions result in performance improvements when we train on interactions and then apply them in the fully inductive setting with no additional interactions done. Finally, protocol 3 shows improvements when we train on the interactions and then do more interactions in the inductive setting. 

The results in this section, combined with the earlier results on Black Lives Matter, show that our approach can generalize across multiple datasets, topics, and events.

\section{Discussion Continued}

\subsection{Model Representations}
\label{discussion:model_representations}

\begin{table}[t]
\begin{center}
\begin{tabular}{|p{3.4cm}|p{3.15cm}|}
  \hline
  {\textbf{\small Model}} & {\textbf{\small Embedding Change \%}}  \\
  \hline
  \small P1: Inductive Human & \small 75.39  \\
  \hline 
  \small P2: Train Human &  \small 64.23 \\
  \hline 
  \small P3: Train + Inter. Human  & \small 51.41  \\
 \hline
\end{tabular}
\end{center}
\vspace{-13pt}
\captionsetup{justification=centering}
\caption{\small Change of node embeddings after interactions compared to the no interaction model on E2-1. Interactions affect model representations (lower $\#$ = more change).}
\vspace{-18pt}
\label{table:representation_change}
\end{table}
Now, in order to measure the impact of interactions on our graph-based model, we evaluate how much model node embeddings change after they are incorporated. To do this, we compute the difference in the cosine similarity of the embedding of each user node before and after interactions are done, and average the results. The results in Tab.~\ref{table:representation_change} show that even a small amount of interactions can make a significant change in model representations. This shows why minimal amounts of interactions can lead to a strong performance increase.

\subsection{Discussion: Interaction Examples}
\label{appendix:human_annotation_analysis}
In this section, we continue our discussion from Sec~\ref{sec:human_annotation_analysis} and provide more examples of nodes that humans connected during the interaction process. We first show the connections humans make (Sec.~\ref{sec:interactions_made}), and then discuss what trends we can learn from these connections about our approach (Sec.~\ref{sec:interactions_trends}). The connections reveal interesting insights on how humans are connecting nodes based on content preferences. Further, it shows that despite all the content being focused on one event, there are lots of different relevant perspectives identified by the model as realistic points of confusion.

\subsubsection{Interactions Made}
\label{sec:interactions_made}

We first show examples of pairs of users/articles that were connected by human interactors, describing what they were about. This analysis was done by the authors based on the human interactions.

\begin{enumerate}
    \item A user with hashtags about taking back the United States by burning and destroying it, and also White Supremacy related hashtags, was connected to an article saying the current President (Biden) was clueless and didn't know what they were doing.
    \item Two users with random and unrelated hashtags in their bio and extremely similar tweet language were connected as they were identified to likely both be bots.
    \item A user that was a sports fan was connected to a source that reported sports media, but in this case had posted an article about how certain races have been negatively impacted from the coronavirus despite being athletic.
    \item An article discussing how the Minnesota Vikings Honored George Floyd's family at their season opener was connected to a source that reported football sports articles that seemed factual.
    \item An atheist, socialist, songwriter, and musician student Twitter user was connected to a Bernie Sanders supporter that wanted student loan forgiveness.
    \item An influencer who was the mayor of a major city was connected to a seemingly politically aligned news reporter for the same city.
\end{enumerate}

Next, we show snippets (to preserve anonymity) of user bios and articles that were connected, to show how simple the process is. We also provide our explanations of why these users/articles were connected. All of these examples are related to the Climate Change event and the text shown is snippets of the actual text that was shown to humans:
\begin{enumerate}
    \item \textbf{Bio 1:} ``what makes you optimistic...sharing optimism of optimistic leaders'' \textbf{Tweet 1:} ``a majority of young people are \#optimistic that it's still possible to prevent the worst effects of \#climatechange''

    \textbf{Bio 2:} ``Christian...\#Goodnews seeker, ther's plenty of it!''

    \textbf{Explanation:} These users were connected by interactors likely because the second user likes good news, and the first user is an optimist specifically sharing good news about climate change!

    \item \textbf{Article 1:} ``...San Diego May Get Climate Update After All..''

    \textbf{Article 2:} ``Fish prices spike as ...face total depletion''

    \textbf{Explanation}: These articles were connected by interactors likely because they both are showing the effects of climate change. It is changing cities, and changing fish prices.

    \item \textbf{Tweet 1:} ``Climate Change...Biggest Hoax in Human History''

    \textbf{Tweet 2}: ``Trump is Hurting Climate Change by letting China take the lead...''

    \textbf{Explanation:} These users were not connected (and so weren't the corresponding articles), and specifically marked \textbf{different}. This is likely because the first user doesn't believe in climate change, while the second is disappointed that President Trump isn't taking more action about it.

    \item \textbf{Article 1:} ``Climate Change...Biggest Hoax in Human History''

    \textbf{Article 2:} ``California bans sale on new gas-powered cars in 2035''

    \textbf{Explanation:} These articles were not connected, and specifically marked \textbf{different}. This is likely because the first doesn't believe in climate change, while the second one does, or at least enough to report on the ban of the sale of gas cars to protect the environment. 
\end{enumerate}

\subsubsection{Interactions Task Details and Trends}
\label{sec:interactions_trends}
While humans can be subjective and make mistakes, we specifically designed our interaction task to be simple to try and eliminate as much of this as possible. Humans were asked to determine user similarity based on how users are discussing certain events, not in depth questions like if a text is factual or not. Determining this high level of user similarity is fairly simple, especially for educated humans, whom we envision performing the interactions.

From these examples above, we can see that our goal to reduce the subjectivity and increase simplicity of our interaction task holds true, at least in our experiments. This is why the entire interaction process can be be done rapidly (humans spent ~3 mins per interaction graph, leading to the creation of 8 edges) and with high human interactor agreement. From the examples shown, it is clear that users/articles were connected based on content match, which was fairly simple for our educated human interactors to tell. However, this is  hard for models, particularly on emerging news events, which is why our interaction setup leads to large performance improvements, even without any training. Also, we note that in most cases, the text defining the user/article similarity was not very subjective, and it is easy to determine the user/article perspective. 

It is also possible, but unlikely, that two users/articles making similar statements don't have at least some similarity on an issue, and thus shouldn't be in the same information community. However, on a large scale over a lot of interactions, the text we show humans is likely to capture user/article perspectives and thus content similarity trends. Thus, even if there are a few rare cases in which users are connected but their statements aren't representative of the community they belong to, it isn't likely to make a significant difference in our learned representation and thus source factuality detection performance.

\end{document}

%% file: introduction.tex
Over the last decade, we have witnessed a rise in the proliferation of ``fake news''~\cite{lazer2018science}, news content which  lacks the journalistic standards ensuring its quality while maintaining its appearance. Social media is flooded with inaccurate and incomplete information~\cite{vosoughi2018spread}, and combating this has attracted significant research interest~\cite{nguyen2020fang}. However, this is still a hard task, particularly on unseen topics. In this paper, rather than annotating data to learn these topics, we propose to use quick \textbf{human interactions} to characterize social media, allowing us to learn a better representation, and detect factuality better.

Instead of fact checking individual articles, some works~\cite{baly:2020:ACL2020} focus on fact-checking sources. While still requiring automated systems due to the number of sources online, source factuality detection can be more scalable, as sources often publish content of similar factuality.  Following this, we focus on capturing the factuality levels of sources: \textit{high, mixed, low}.

One concept underlying methods that aim to exploit social information for identifying the factuality of news sources is the \textit{social homophily principle}~\cite{mcpherson2001birds}, which captures the tendency of members of the same social group to hold similar views and content preferences. This often leads to the formation of  \textit{``echo chambers''}~\cite{jamieson2008echo,quattrociocchi2016echo}, tightly-knit \textit{``information communities''} that have little interaction with other communities holding different views. Prior work shows how similar news, particularly misinformation, tends to spread more in some of these tightly-knit information communities \cite{bessi2016homophily}. Thus, identifying them can provide the needed information for capturing the factuality of sources (communities spreading mostly low factuality content in the past are likely to spread low factuality content in the future). In this work, we first capture social information in an information graph, modeling it via a R-GCN~\cite{schlichtkrull2018modeling}.

Many approaches to detect news factuality are often studied in unrealistic settings, as their success hinges on test data being similar to or related to training data. However, a more realistic setup would examine whether a system would be able to generalize to emerging news events: These events introduce different narratives, users, and news sources, that are unseen and do not interact with training content; i.e. test users don't follow train users and test graph nodes aren't connected to train nodes. In this paper, to simulate these settings, we collected new data, consisting of the articles published around specific news events (Black Lives Matter and Climate Change - see Sec~\ref{sec:dataset}), their sources, and social context. We applied a recent strong baseline system~\cite{mehta2022tackling}, trained over data sampled from past events, and it resulted in significant degradation in performance on the new events ($\sim$22\% Acc, 19\% Macro-F1).

\begin{figure*}[t!]
  \centering
  \includegraphics[scale=0.35]{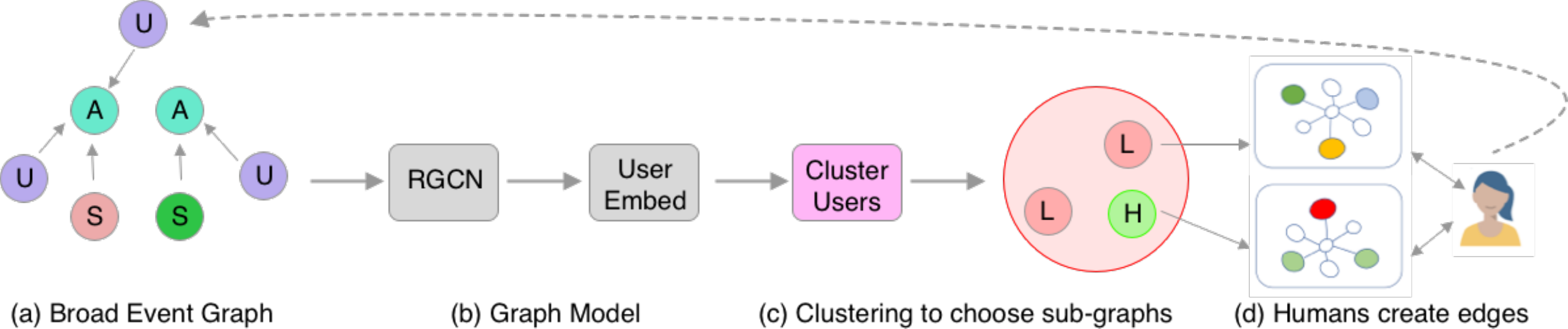}
\caption{\small Our framework overview: \textbf{Adapting News Source Factuality Detection to Emerging News Events by Interactively Characterizing the Social Media Context of News Articles and Their Sources}. (Key: U = Users, A = Articles, S = Sources, Green/H = High Factuality, Red/L = Low Factuality). From the learned graph model (b), we find pairs of inconsistent users by clustering all user embeddings and looking for conflicting factuality labels (c) (L = low; H = high factuality). Here, the High Factuality user doesn't match the mostly Low Factuality cluster. We then build sub-graphs from these pairs of mismatched users and their community to show human interactors (d), who create new edges based on content similarity. This is far simpler and quicker than identifying factuality, as humans only need to identify which nodes are similar in content. Based on the interactions, we create edges in the broad event graph (a) to do better news source factuality detection (either directly or with more training).}
\label{fig:example-intro}
\vspace{-10pt}
\end{figure*}

Our main observation in this paper is that even in these challenging settings, the social homophily principle can be exploited to better detect source factuality, \textbf{if the system can identify relevant information communities over users engaging with the new content}. This is since users that are part of an information community that propagates fake news, are likely to do so as well. As we later show, automatically detecting the factuality of news sources is difficult, particularly on emerging news events. \textit{Instead}, we suggest an \textbf{interactive learning protocol}, in which human judgements dynamically help the model identify these communities. As humans analyzing all emerging news content is clearly infeasible, we propose a novel sampling method for interactions, based on resolving inconsistencies in the model’s graph-based social representation. Specifically, we identify pairs of users that are clustered in the same community, but have conflicting factuality predictions, as this indicates inconsistency. We create small sub-graphs corresponding to the social and content preferences of these users and other members of the community, and ask the humans to resolve the conflict: Based on their profile descriptions, social relations and articles endorsed, \textit{is it likely (given the principle of social homophily) that these two users belong to the same community?} The human judgements provide rich feedback for this question, by adding edges to the graph, which connect users, articles, and sources. These edges result in cleaner information communities, which alleviate the difficulty of the source classification task. Fig.~\ref{fig:example-intro} describes this.  

In summary, we make the following contributions. (1) We are the first to formulate the task of \textbf{interactive news source factuality detection} by characterizing social context, and implement an interaction tool for supporting this. (2) We suggest a novel sampling approach for reducing the number of human judgements needed by focusing on social inconsistencies.  (3) We focus on one of the most challenging settings of news source factuality detection in emerging news events, collect data, and perform experiments showing how minimal, quick interactions can lead to performance improvements on unseen data. More generally, we propose an interactive framework to learn stronger information communities, and apply it to improve news source factuality detection. In the future, it can also be applied to other social media analysis tasks. 

Sec.~\ref{sec:model} describes our graph model, Sec.~\ref{sec:interactions} our novel protocol to incorporate interactions, Sec.~\ref{sec:experiments} shows results, and Sec.~\ref{sec:discussion} analyzes them.

%% file: related.tex
Detecting fake news on social media is a popular research topic, studied in supervised learning ~\cite{hassan2017toward,perez2017automatic,volkova2018misleading, ma2018rumor, shu2019defend, shu2019fakenewstracker, kim2019homogeneity}, Graphs \cite{han2020graph, li2022dynamic},  zero-shot \cite{wright2022generating}, dialogue \cite{gupta2022dialfact}, cross-domain \cite{huang2021dafd, ENDEF, zhu2022memory, mosallanezhad2022domain}, and low-resource \cite{lin2022detect} settings.

One of the most challenging yet most critical social context fake news detection settings is the early detection of it, where test data has new users, articles, and sources, that do not interact with training data. Recently, researchers have been working on this task, especially at the article/tweet level. \citeauthor{liu2018early} classify news propagation paths, \citeauthor{yuan2020early} model user credibility, while \citeauthor{konkobo2020deep} built a semi-supervised classifier. In our work, we focus on this challenging early detection setting, specifically to identify the factuality of news sources. We show how our \textit{interactive setup} can be useful, even in these settings. If combined with other early detection methods, our framework may lead to further gains, and we leave this for future work.

Using human interactions to improve models has also been popular recently~\cite{internlp-2021-interactive}, in scenarios such as active learning \cite{blok2021active},  or humans providing general system feedback \cite{tandon2022learning}. Other works exploit human feedback for concept discovery~\cite{pacheco-etal-2022-holistic,pacheco-etal-2023-interactive} by communicating human-level symbolic knowledge~\cite{pacheco2021modeling}. In contrast, our interactions enable stronger general models, and generalization to new unseen scenarios.

Social homophily has been used to better many NLP tasks, like sentiment analysis, entity linking, and fake news. \cite{west2014exploiting, yang2016toward, mehta2022tackling}. Particularly, prior work shows how misinformation (and similar news) spreads more in tightly-knit communities, motivating our idea that if we use humans to increase homophily and build better information communities, we can detect facutality better \cite{bessi2016homophily, halberstam2016homophily, cinelli2021echo}.

%% file: model.tex
Similar to \citeauthor{mehta2022tackling}, we view fake news source detection as reasoning over relationships between sources, articles, and users in an information graph. We use their graph model\footnote{\url{https://github.com/hockeybro12/FakeNews_Inference_Operators}}, briefly explaining it in this sec. Sec.~\ref{sec:interactions} explains our interactive protocol.

The model uses a heterogeneous graph to capture the interaction between social information and news content, and a Relational Graph Convolutional Network (R-GCN) to encode it. The R-GCN allows us to create contextualized node representations for factuality prediction. For example, one way sources are represented is by the articles they publish (which in turn are also represented by their relationships to other nodes).

\textbf{Graph Creation}: The graph (see: Fig.\ref{fig:example-intro}a) consists of 3 types of nodes, each with feature vectors (details: App.~\ref{subsec:initial_embeddings}): (1) $S$, the news \textit{sources}, are our classification targets. 
(2) $A$, the \textit{articles} published by these sources,
(3) $U$, the Twitter \textit{users}. Sources are first connected to articles they publish. Social context is added via Twitter users that interact/connect to sources/ articles/other users. These users provide the means for fake (and real) news spread on social media: \textbf{(1) Following Sources/Users:} Users are connected to sources and users they follow. \textbf{(2) Propagating Articles:} Articles are connected to users that tweet its title/link.

\textbf{Graph Embedding:} As in \citeauthor{mehta2022tackling}, we train a R-GCN \cite{schlichtkrull2018modeling} to learn graph embeddings, which will be later used to determine where human interaction may be beneficial. We optimize the Classification objective of News Source Factuality Detection (categorical cross-entropy). To predict labels, we pass the source node embeddings from R-GCN through the Softmax activation.

%% file: interactions.tex
We hypothesize that understanding content and the context it is provided in is critical to detecting fake news. Specifically, identifying information communities of users, sources, and articles based on their content preferences can be helpful, as a community that mostly shares fake news in the past, is likely to share fake news in the future. Further, users that join this community are likely to share beliefs of the community, and thus also share fake news.

Unfortunately, understanding content on social media and using it to identify information communities is challenging for AI agents. It becomes more difficult as new events with new relationships arise, as the agent does not have enough data to determine what is fake news. This makes the early detection of fake news difficult (see Sec.~\ref{sec:experiments_baselines}). On the other hand, educated humans can more easily understand relationships on social media, even in new events, as they can better analyze social interactions. Thus, humans can clear up model confusion by helping the model identify the information communities or make existing ones bigger. For example, after reading a sample of tweets from users discussing a new event, humans can \textit{quickly} determine if the users are offering the same perspectives, and should be in the same community. This knowledge can help the agent model these users and other content they interact with better. As we later show experimentally, human interactions like these enables us to build strong information communities, which helps the agent, particularly with the early detection of news sources factuality on new news events. 

Unlike automated agents, humans cannot analyze all content that pertains to a new event, as it is too massive. Instead, due to the highly connected structure of social media, \textbf{small amounts of interactions done in the right places can make significant impact}, as the added information can flow throughout the information graph. Thus, we first discuss in Sec~\ref{sec:getting_human_interactions} how we determine what content humans should interact with and what interactions they should make (i.e. forming/strengthening information communities). Then, in Sec~\ref{sec:incorporating_human_interactions}, we explain how we can incorporate those interactions back into the model to achieve performance improvements.

\subsection{Soliciting Human Interactions}
\label{sec:getting_human_interactions}

Now, we discuss 3 different protocols to identify the data on which humans should interact, and then what they should do. In general, we want humans to analyze a sub-graph of the broad information graph characterizing the new event. Given this sub-graph, we ask humans to help form information communities by characterizing the content in the graph based on similarity, i.e. identify if two users are similar, two articles offer the same perspective, etc. 
This is done by asking humans a series of questions (details : App.~\ref{appendix:supplemental_advice_graphs}) which enables them to connect nodes in the sub-graph based on content preferences, via a graphical interface we developed. An ex. is shown in App. Fig~\ref{fig:interaction_graph_example}. We then replicate these connections in the broad information graph.

Identifying the sub-graph that will benefit the most from interactions is critical to getting the most value out of each interaction. We build the sub-graphs by first choosing a pair of users, as our end goal is to build stronger user information communities. We explore three different protocols for doing this in \ref{subsection:baselines_interactions} and Sec~\ref{subsection_social_factuality_mismatch}. After finding these pairs of users, we build the sub-graph to show humans by including these users and their direct connections in the graph. This includes the articles they propagate, other users that propagate those articles, the sources that publish those articles, and up to 3 ``influencers'' (users with over 1000 followers) that one of these users follows. For each node in the sub-graph, we populate it with relevant information to enable the human interactors to understand content. For ex:, user/source nodes show user bio, tweets, etc. Article nodes show article publish date, headline, and first paragraph. Details: App~\ref{appendix:supplemental_advice_graphs}.

\subsubsection{Baselines}
\label{subsection:baselines_interactions}
We have two baselines for selecting pairs of users. \textbf{(1) Random}, users at random.  \textbf{(2) Model Confusion} takes an \textit{active learning}-like approach, and chooses users based on a label confusion criterion, calculated by propagating the softmax score of the source prediction downwards to get user confusion. Specifically, to get this score, we look at all the sources the user directly interacts with (articles they propagate and sources they follow), and then take the weighted average of those source's Softmax predicted label to be the user score (thus approximating user confidence). For example, a user interacting with 3 articles predicted with low factuality score of 0.7 and 1 source with high factuality score 0.9 will have confidence 0.75.

\subsubsection{Social vs. Factuality Mismatch Criterion}
\label{subsection_social_factuality_mismatch}

Now, we discuss our novel protocol to determine the pairs of users, seen also in Alg~\ref{alg:soliciting_interactions}. It is designed around one of the key ideas in this paper, homophily, the tendency of users with similar social preferences to have similar content preferences. Our graph model learns to represent both, by creating node embeddings which capture users' similarities, and learning classifiers used for characterizing content by identifying factuality. Intuitively, our protocol is designed to identify users, that based on the current model parameters, break the homophily principle. These users are part of the same social group while at the same time have different factuality predictions, and thus likely different content preferences. When this is true, the model may not have clearly understood the content preferences of these users, which a human can help clear up.

To identify these pairs of users, we first need to compute factuality labels for each user. As the model is trained for source classification, we designed a heuristic to use source labels to compute user labels: We assign users the label of the \textit{most common predicted label} of the sources/articles the user is directly connected to. For ex:, a user following 3 low factuality sources and tweeting 1 mixed factuality article is assigned a low factuality label, as it interacts with more low factuality content.

After computing user labels, we need to find groups of similar users, which we do by k-means clustering all users in the event graph using their model embeddings (Alg~\ref{alg:soliciting_interactions}: 3). Then, we assign each cluster a factuality label based on the most frequently occurring user factuality label in that cluster (Alg~\ref{alg:soliciting_interactions}: 5). Finally, we choose pairs of users that are in the same cluster, but one has a different label than the cluster label, as the model thinks they are similar but predicts their factuality differently, which indicates a sign of confusion (Alg~\ref{alg:soliciting_interactions}: 7-9).

\begin{algorithm}
\caption{\textit{Social vs. Factuality Mismatch}}
\begin{algorithmic}[1]
  \small
  \STATE \textbf{Input:} $U$ (Users), $U_E$ (Graph User Embeddings), $F$ (User Factuality Scores), $P$ (Empty List)
  \STATE \textbf{Output:} $P$ (Pairs of Users To Build Graphs)

  \STATE $c_{1...k} = \text{k-means}(U_E)$ {\scriptsize{\texttt{K-means Cluster all Users based on Graph User Embeddings}}}

  \FORALL[for each cluster]{$i=1, \dots, k$}
  \STATE $c_l = \max_{0 \leq u \leq n} F_u$ {\scriptsize{\texttt{Assign Cluster the Label of the most common user}}}
  \FORALL[for each user]{$j=1, \dots, n$}
  \IF[If user label $\neq$ cluster label]{$F_j \neq c_l$}
  \STATE $U_k = \text{rand}(U)$ \text{ where } ($F_j \neq F_k) \land (F_k \in c_l)$  {\scriptsize{\texttt{Choose a random user in the cluster with a different label}}}
  \STATE $P.\text{insert}((U_j, U_k))$ {\scriptsize{\texttt{Add User Pair to List}}}
  \ENDIF
  \ENDFOR
  \ENDFOR

  \RETURN $P$ (Pairs of Users)
\end{algorithmic}
\label{alg:soliciting_interactions}
\end{algorithm}

\subsection{Incorporating Human Interactions}
\label{sec:incorporating_human_interactions}
Humans interact by making new connections on the sub-graphs. We then utilize the interactions by connecting the appropriate nodes in the broader event graph. Our goal is to show how human interactions allow us to have a better model that performs well with and without further interactions. 

We focus on the challenging \textbf{fully inductive setting}: where all test set nodes are not seen at training and are also not connected to training set nodes. Further, we evaluate the important setting of early detection of fake news, where test data comes from unseen emerging events. As we show in Sec~\ref{sec:experiments_baselines}, in these settings, existing models struggle. 

We evaluate 3 interaction-based protocols. The three protocols have the same starting point, a graph-based factuality classification system trained over an established dataset~\cite{baly:2020:ACL2020}. The protocols are designed to show how interactions can enhance that initial system when making predictions on data from unseen emerging events, and are organized in order of increasing effort required and increasing performance.
All involve performing interactions on up to two different data sets (each corresponding to a different emerging event, see Sec~\ref{sec:dataset}). Since some of the protocols we introduce update the parameters of the model after interaction, we collect data for two events to ensure that all protocols can be evaluated in the fully inductive settings on the second event data (i.e., relying on interactions alone without training).

We hereby refer to the first event as $E1$, and the second as $E2$. Each event is further split into interaction and no interaction halves (ex: $E1$-$1 / E1$-$2$), for comparison and model training (see below).

\textbf{(1) Fully Inductive:} In the first protocol, humans interact on the interaction halves of $E1$ and $E2$, and then the interactions are incorporated, without any additional training. This is the most challenging, but no extra effort is necessary for performance improvements.

\textbf{(2) Interactions Amplify Model Learning:} Here, our goal is to show how interactions can help us learn a stronger model that performs well without interactions. Thus, we interact on the interaction half of $E1$ (half so we can evaluate how we do on the same event without interactions), use it to train the model, and evaluate it on $E2$ (future event, fully inductive) without any additional interactions.

\textbf{(3) Learning to Incorporate Interactions:} In this protocol, we show how training the model after interactions allows the model to learn how to better incorporate them. This enables it to do even better when interactions are provided on future events. To do this, as above, we interact on half of $E1$ and train on it. Then, we evaluate $E2$ on both the interaction and non-interaction half. Both halves of $E2$ are connected, so although interactions are only on half, information can propagate via the graph.

\subsection{Simulating Human Interactions}
\label{sec:simulating_human_interactions}
Due to constraints involved with human interaction time/cost, to evaluate our models we also designed a heuristic to simulate humans: We hypothesize that two users are similar if they have the same gold factuality label. While our interaction approach prioritizes content preferences for interactions, identifying this automatically is difficult, so this is an approximation. Thus, doing human interactions at the scale of simulated ones could perform better, and we leave it for future work. To get user gold factuality labels, we use the same heuristic as in Sec~\ref{sec:getting_human_interactions} (assinging users the label of the source they are most often connected to). Note that in this simulated interaction setting, we are using the test set to determine user labels, so this setup is not realistic.

%% file: experiments.tex
\subsection{Dataset and Collection}
\label{sec:dataset}
\input{data.tex}

\subsection{Evaluation Method}
\label{sec:evaluation_method}
For both BLM and CLM, we evaluate on the two inductive sub-events ($E1, E2$) collected in Sec~\ref{sec:dataset}. The interaction half is referred to with a -$1$ and the non-interaction with -$2$, for ex: $E1$-$1$. For fair comparison, each data split, i.e. $E1$-$1$, is the same across all evaluations. We report results on Accuracy, Macro F1 (the dataset is unbalanced), and the total number of edges added by all interactions.

We evaluated 3 settings, the first 2 are simulated using gold test set labels (see Sec~\ref{sec:simulating_human_interactions}), while the last is done by humans: \textbf{(1) Interaction Graphs Only:} Edges are added only between users in interaction graphs. \textbf{(2) X\% of Data:} Edges are added between X\% of all possible users that have the same label in the test set that we run interactions on (not only users in interaction sub-graphs). X is 100\%, 75\%, or 25\%. \textbf{(3) Human Interactions}: For BLM, we evaluate on two separate versions, each featuring a different source set (and social media data). This section shows results on the first version, when one human interacts on 20 graphs per data split (details in \ref{appendix:interaction_graph_details}, \ref{appendix_human_interaction_details}). The appendix shows the second version of BLM with \textbf{\textit{3 different interactors}}, showing the same trends. For space, these interaction results/details (including agreement) are in \ref{sec:additional_interactor_details}. This section also evaluates Climate Change, with 2 interactors interacting on 10 sub-graphs per data split. (for detailed CLM results, see App. \ref{sec:climate_change_results}). The human interaction results are also the most realistic evaluation setting, as they don't use any gold test set labels, like the simulated interactions do.

\subsection{Baselines}
\label{sec:experiments_baselines}
We trained our baseline model, from Sec.~\ref{sec:model}, for Source Factuality Detection on \citeauthor{baly:2020:ACL2020} and the first event, where it achieved strong performance, similar to SOTA \cite{mehta2022tackling} (we use the same data and methodology) and other baselines \cite{baly:2020:ACL2020} (SVM). However, when evaluated inductively on a BLM event that was published after dates the training data was collected from - i.e. $E1$-$1$ - performance significantly worsened (see Table~\ref{table:baseline_results}). This validated our hypothesis that strong models, even if trained on generic and event specific data, do not translate well to future events. Thus, we propose to use our interactive protocol.

\begin{table}
\begin{center}
\begin{tabular}{|p{2cm}|p{1.5cm}|p{1.5cm}|}
  \hline
  {\textbf{\small Model}} & {\textbf{\small  E2-1 Acc}} & {\textbf{\small  E2-1 F1}} \\

 \hline
  \small  Random Users  & \small 35.21 & \small 29.99 \\
  \small  Confused Users  & \small 36.61 & \small \textbf{32.72} \\
  \small  User Clustering  & \textbf{\small 42.10}  & \small 32.22  \\
  
 \hline
\end{tabular}
\end{center}
\vspace{-10pt}
\captionsetup{justification=centering}
\caption{\small Ablation study on our methods for choosing interactions on $E2$-$1$. It is clear that finding users based on clustering and then factuality mismatch is best.}
\vspace{-15pt}
\label{table:soliciting_interactions}
\end{table}

\begin{table*}
\begin{center}
\begin{tabular}{|p{6.9cm}|p{0.8cm}|p{0.8cm}|p{0.8cm}|p{0.8cm}|p{0.8cm}|p{0.8cm}|p{0.8cm}|}
  \hline
  {\textbf{\small Model}} & {\textbf{\small E1-1 Acc}}
  & {\textbf{\small E1-1 F1}} & {\textbf{\small E1-2 Acc}} & {\textbf{\small E1-2 F1}} & {\textbf{\small E2-1 Acc}} &  {\textbf{\small E2-1 F1}} & {\textbf{\small \# Edges}} \\

 \hline
  \small BLM No Interactions & \small 43.21 & \small 34.44 & \small 37.93 & \small 30.70 & \small 35.21 & \small 27.65 & \small - \\
  \small CLM No Interactions & \small 40.16 & \small 32.77 & \small 39.65 & \small 31.86 & \small 34.88 & \small 30.93 & \small - \\
  \hline
  \small BLM Sim. Interactions on Sub-Graphs Only & \small 44.54 & \small 36.45 & \small 37.93 & \small 30.70 & \small 42.10 & \small 32.22 & \small 2,162 \\
  \small BLM Sim. Interactions on 100\% of Data in E1-1 + E2-1 & \small 49.20 & \small 40.52 &  \small 37.93 & \small 30.70 & \small 44.73 & \small 36.82 &  \small 133,336 \\
  \small BLM Sim. Interactions on 75\% of Data in E1-1 + E2-1 & \small 46.03 & \small 38.05 & \small 37.93 & \small 30.70 & \small 50.00 & \small 40.50 & \small 74,414 \\
  \small BLM Sim. Interactions on 25\% of Data in E1-1 + E1-2 & \small 46.03 & \small 37.65 & \small 37.93 & \small 30.70 & \small 42.10 & \small 32.65 & \small 8,266 \\
  \hline 
  \small BLM Human Interactions in E1-1 + E2-1 & \small 44.44 & \small 35.96 & \small 37.93 & \small 30.70 & \small 44.73 & \small 30.03 & \small 84 \\
  \small CLM Human Interactions in E1-1 + E2-1 & \small 46.72 & \small 43.94 & \small 39.65 & \small 31.86 & \small 39.53 & \small 36.95 & \small 47 \\
  
 \hline
\end{tabular}
\end{center}
\vspace{-10pt}
\captionsetup{justification=centering}
\caption{\small Protocol 1: Interactions results on BLM and Climate Change (CLM) in the difficult, inductive, no training setting. E1 and E2 are the two separate, inductive graphs. E1-1 is the first half that receives interactions, and E1-2 is the second half that doesn't. E2-1 (first half E2) also receives interactions, but it's dev set is not used to select the model. With a minimal number of added edges, human interactions achieve performance improvements in these difficult, inductive settings, with no extra training (compared to No Interactions). Ex: results improve on human BLM E2-1 ($\sim$9.5\% Acc.) Sim. settings also show improvements.}
\label{table:use_case_1_inductive}
\end{table*}

\begin{table*}[t]
\begin{center}
\begin{tabular}{|p{5.99cm}|p{0.9cm}|p{0.9cm}|p{0.9cm}|p{0.9cm}|p{0.9cm}|}
  \hline
  {\textbf{\small Model}} & {\textbf{\small E1-2 Acc}} & {\small \textbf{E1-2 F1}} & {\small \textbf{E2-1 Acc}} &  {\small \textbf{E2-1 F1}} & {\small \textbf{\# Edges}} \\

 \hline
  \small BLM No Interactions & \small 37.93 & \small 30.70 & \small 35.21 & \small 27.65 & \small - \\
  \small BLM No Interactions Train & \small 64.86 & \small 66.91 & \small 42.10 & \small 40.10 & - \\
  \small CLM No Interactions Train & \small 49.29 & \small 44.84 & \small 44.77 & \small 42.35 & - \\
  \hline
  \small BLM Sim. Interactions on Sub-Graphs Only & \small 62.16 & \small 62.95 & \small 43.66 & \small 29.47 & \small  2,162\\
  \small BLM Sim. Interactions on 100\% of Data in E1-1 & \small 56.75 & \small 59.27 & \small 45.07 & \small 40.18  &   \small 133,336 \\
  \small BLM Sim. Interactions on 75\% of Data in E1-1 & \small 65.51 & \small 64.01 & \small 43.66 & \small 39.11 & \small 74,414 \\
  \small BLM Sim. Interactions on 25\% of Data in E1-1  & \small 54.05 &  \small 46.48 & \small 39.43 &  \small 35.09 & \small 8,266 \\
  \hline 
  \small BLM Human Interactions in E1-1 & \small 67.56 & \small 71.56 & \small 45.07 & \small 35.18 & \small 84 \\
  \small CLM Human Interactions in E1-1 & \small 53.52 & \small 44.53 & \small 40.29 & \small 46.38 & \small 47 \\
  
 \hline
\end{tabular}
\end{center}
\vspace{-10pt}
\captionsetup{justification=centering}
\caption{\small Protocol 2: Interactions results on BLM + Climate Change (CLM) when we train on interactions, and then apply the model to a new event with no interactions done. E1 and E2 are the two separate, inductive graphs. E1-1 is the interaction half of the $1^{st}$ event and E1-2 is the $2^{nd}$, non-interaction half. E2-1 (non-interaction half) is not connected to E1. Compared to the model that was trained on E1-1 without interactions (No Interactions Train), human interactions lead to a more accurate model for future events, by $\sim$3\% better Acc. for BLM and $\sim$4\% F1 for CLM (E2-1). Sim. settings also show improvements.}
\label{table:use_case_2_train}
\vspace{-15pt}
\end{table*}

\subsection{Interactions}
\label{sec:experiments_simulated_advice}
We now evaluate our interaction protocols - what portions of the graph to show users and how to incorporate interactions, using the method in Sec~\ref{sec:evaluation_method}.

\subsubsection{Soliciting Interactions}
\label{sec:experiments_soliciting_interactions}

When comparing our methods for choosing what sub-graphs to show on BLM, simulated interactions performance shows a benefit (Table~\ref{table:soliciting_interactions}) of choosing the users to build interaction graphs for based on confused user clustering. This matches our intuition as if the model predicts a users' factuality differently than other users similar to it, then the model is confused and clearing that could improve performance. Thus, we use this method of choosing sub-graphs throughout the rest of our experiments.

\begin{table*}[t]
\begin{center}
\begin{tabular}{|p{6.9cm}|p{0.8cm}|p{0.8cm}|p{0.8cm}|p{0.8cm}|p{0.8cm}|p{0.8cm}|p{0.8cm}|}
  \hline
  {\textbf{\small Model}} & {\textbf{\small E1-2 Acc}} & {\textbf{\small E1-2 F1}} & {\textbf{\small E2-1 Acc}} &  {\textbf{\small E2-1 F1}} & {\textbf{\small E2-2 Acc}} &  {\textbf{\small E2-2 F1}} & {\textbf{\small \# Edges}} \\

 \hline
  \small BLM No Interactions & \small 37.93 & \small 30.70 & \small 35.21 & \small 27.65 & \small 30.30 & \small 24.84 & \small - \\
  \small BLM No Interactions Train & \small 64.86 & \small 66.91 & \small 42.10 & \small 40.10 & \small 45.45 & \small 42.35 & \small - \\
  \small CLM No Interactions Train & \small 49.29  & \small 44.84 & \small 44.77 & \small 42.35 & \small 44.44 & \small 33.06 & \small - \\
  \hline
  \small BLM Sim. Interactions on Sub-Graphs Only  & \small 62.16 & \small 62.95 & \small 57.89 & \small 48.53 & \small 45.45 & \small 43.49 & \small 2,162 \\
   \small BLM Sim. Interactions on 100\% of Data in E1-1 + E1-2 & \small 56.75 & \small 59.27  & \small 57.89 & \small 61.90 & \small 36.36 & \small 35.53 & \small 133,336 \\
  \small BLM Sim. Interactions on 75\% of Data in E1-1 + E1-2  & \small 65.51 & \small 64.01 & \small 63.15 & \small 61.84 & \small 45.45 & \small 43.60 & \small 74,414 \\
  \small BLM Sim. Interactions on 25\% of Data in E1-1 + E1-2  & \small 54.05 & \small 46.48 & \small 44.73 &\small  31.38 & \small 51.51 & \small 45.31 & \small 8,266 \\
  \hline 
  \small BLM Human Interactions in E1-1 + E2-1  & \small 67.56 & \small 71.56 & \small 50.00 & \small 43.60 & \small 51.51 & \small 40.09 & \small 84 \\
  \small CLM Human Interactions in E1-1 + E2-1  & \small 53.52 & \small 44.53 & \small 53.48 & \small 43.07 & \small 46.80 & \small 38.73 & \small 47 \\

 \hline
\end{tabular}
\end{center}
\vspace{-15pt}
\captionsetup{justification=centering}
\caption{\small Protocol 3: Results on BLM + Climate Change (CLM) when we train on interactions and then do more in the inductive setting. E1 and E2 are the two separate inductive graphs. E1-1 is the interaction half of E1 that is trained on. E1-2 is the non-interaction half. E2 receives interactions on the interaction half (E2-1), but not the non-interaction half (E2-2). Human interactions improve accuracy on both halves of E2 and F1 on E2-1, compared to no interactions train, and more than only applying interactions without training for them as Tab.\ref{table:use_case_1_inductive}, showing the benefit of training to learn to incorporate interactions.}
\label{table:use_case_3_train_learn_advice}
\vspace{-10pt}
\end{table*}

\subsubsection{Incorporating Interactions}
\label{sec:experiments_incorporating_interactions}
Now, we evaluate our 3 protocols of incorporating interactions discussed in Sec~\ref{sec:incorporating_human_interactions}, in order of increasing performance and model training required. For space, additional human interaction results are in App.~\ref{sec:additional_interactor_details} and detailed CLM results in App.~\ref{sec:climate_change_results}. Note that simulated interactions (\ref{sec:simulating_human_interactions}) use gold test set labels and thus are only used to test our models.

First, Protocol 1, where we evaluate how the model performs with interactions in the completely inductive setting, so no training is necessary. In Tab.~\ref{table:use_case_1_inductive}, we ran interactions on only the interaction half of each event ($E1$-$1$ + $E2$-$1$) and the dev. data (also from $E1$-$1$), to choose the strongest model. To ensure the dev. set being chosen from $E1$-$1$ does not bias us into a strong model, we also did interactions on the interaction half of $E2$ and notice stronger performance improvements. Note that $E2$ is a future event and is not connected to $E1$ at all. All settings improve performance. Moreover, on BLM, human interactions improves performance $\sim$9.3\% Acc. on E2-1, comparable to simulated interactions with significantly more data, showing the large impact benefit of human interactions. 

Next, in Protocol 2, we learn a better model for news source factuality detection after doing interactions, compared to not doing any. In Tab.~\ref{table:use_case_2_train}, we ran interactions on the interaction half of $E1$, and then trained on that data. On $E2$ with no interactions done, we can see how this improves accuracy compared to models trained without interactions. 

Finally, for Protocol 3, we learn to better incorporate interactions into the model after we train for it. Thus, we train similarly to Protocol 2, but now we also run interactions on the interaction half of $E2$. In Tab.~\ref{table:use_case_3_train_learn_advice}, we see accuracy improves on both halves of $E2$ after we learn to incorporate interactions on $E1$, even though $E2$ is inductive. Further, F1 improves on $E1$-$1$. This shows that training with and then doing interactions helps performance significantly on future events. We hypothesize that this happens as training with interactions enables the model to learn how to incorporate them better, allowing the model to further take advantage of them whenever provided. Further, human interactions based on content preferences provide clearer results compared to simulated ones (without cheating and using test set labels), as the model better learns the social media landscape, shown by it achieving better accuracy on the BLM interacted and non-interacted data (both halves of $E2$).

From these results, we see that our real-world applicable human interactions models result in performance improvements in either Accuracy or Macro F1, often times both. As a whole, all our models improve performance (any non-gain in one of these metrics is offset by significant gains in the other). We additionally hypothesize that performing more interactions (particularly human) will achieve higher and more consistent results.

%% file: data.tex
To evaluate our model's ability to predict the factuality of news medium, we used the Media Bias/Fact Check dataset \cite{baly:2018:EMNLP2018}. We expand it by scraping additional sources from Media Bias/Fact Check\footnote{https://mediabiasfactcheck.com}, for better coverage of recent events and increasing the number of sources for evaluation. Identically to \citeauthor{baly:2018:EMNLP2018}, we labeled the sources on a 3-point factuality scale: \textit{high}, \textit{mixed}, or \textit{low}. 

Our goal in this paper is to show how human interactions can help news source factuality detection on new events, where even strong models struggle (Sec~\ref{sec:experiments_baselines}). To do this, we evaluated our model on \textbf{two broad events}: \textit{Black Lives Matter (BLM)} and \textit{Climate Change (CLM)}. For each, we scraped data from Twitter over 3 time periods (01/02/2019 -  06/01/19; 06/02/19 - 01/1/21; 02/02/21 - 05/06/22), each of which additionally cover many different sub-events. For each time period, we created a fully inductive graph, consisting of at least 99 sources and their metadata. None of these graphs are connected to each other in any way and no nodes in any of them are common with each other or the training set - making our test settings fully inductive, and very challenging. To ensure this inductive setting, when collecting data for future time periods, we made sure not to include sources/users/articles that we already used in previous time periods, even if they propagated content in those future periods. Combined with \citeauthor{baly:2020:ACL2020}, we used the first time period for training, to teach the model how to identify fake news in general and as it pertains to an event. We used the 2nd and 3rd time period as $E1$ and $E2$ in the protocols discussed in Sec~\ref{sec:incorporating_human_interactions}. Details (statistics, etc.): \ref{sec:event_collection}. We release our code and data.\footnote{\url{https://github.com/hockeybro12/Fake_News_Interactive_Detection}}

\begin{table}
\begin{center}
\begin{tabular}{|p{2.4cm}|p{0.7cm}|p{0.7cm}|p{0.7cm}|p{0.7cm}|p{0.7cm}|p{0.7cm}|p{0.7cm}|}
  \hline
  {\textbf{\small Model}} & {\textbf{\small Baly Acc.}}
  & {\textbf{\small Baly F1}} & {\textbf{\small E1-1 Acc}} & {\textbf{\small E1-1 F1}} \\

 \hline
  \small Baly & \small 71.52 & \small 67.25 & - & - \\
  \small Mehta R-GCN & \small 68.90 & \small 63.72 & - & - \\
  \small Mehta BEST & \small 72.55 & \small 66.89 & - & - \\
  \hline
  \small BL: Mehta R-GCN & \small 66.04 & \small 54.20 & \small 43.21 & \small 34.44 \\
  
 \hline
\end{tabular}
\end{center}
\vspace{-10pt}
\captionsetup{justification=centering}
\caption{\small Baseline results on Baly \cite{baly:2020:ACL2020} and an inductive future BLM event $E1$-$1$ (not seen or connected to the training graph). Baseline (BL) is the strong graph classification model from \cite{mehta2022tackling} (Mehta R-GCN) that was competitive with the state of the art (\cite{mehta2022tackling} - Mehta BEST). Even with this, performance significantly worsens on $E1$, showing that detecting fake news on future events inductively is challenging. BL: Mehta R-GCN was trained on a smaller \cite{baly:2020:ACL2020} dataset, as some sources were used for evaluation, which is why the performance is slightly lower.}
\vspace{-15pt}
\label{table:baseline_results}
\end{table}

%% file: discussion.tex
Now, we analyze our best BLM interaction model for fake news source detection (for each protocol) on $E2$-$1$ by answering these research questions:\newline
    (1) \textit{Do interactions help learn better communities?} \newline
    (2) \textit{What pairs of nodes do humans connect?}\newline
    (3) \textit{How can our model be used in the real world?} \newline
    (4) \textit{Do interactions change embeddings?} App.~\ref{discussion:model_representations}
\subsection{Learned Communities}

\begin{table}[t]
\begin{center}
\begin{tabular}{|p{3.4cm}|p{2cm}|p{1cm}|}
  \hline
  {\textbf{\small Model}} & {\textbf{\small Purity}} & {\textbf{\small \# Edge}} \\

 \hline
  \small No Inter. & \small 36.2, 37.8, 33.3  &  \small - \\
  \hline
  \small P1: Inductive Human & \small 39.2, 40.2, 35.3 &  \small 84 \\
  \hline 
   \small P2: Train Human & \small 49.5, 37.4, 41.4  &  \small 84 \\
  \hline 
   \small P3: Train + Inter. Human & \small 53.4, 41.9, 42.6 &  \small 84 \\
 \hline
\end{tabular}
\end{center}
\vspace{-10pt}
\captionsetup{justification=centering}
\caption{\small Purity clustering (sources, articles, users) for the human interaction protocols on E2-1. As training increases with each protocol (P), purity does too, showing that interactions do help to learn better information communities.}
\vspace{-16pt}
\label{table:cluster_purity}
\end{table}

We analyze how interactions help learn better info. communities. We evaluate cluster-purity by K-means clustering sources, articles, and users before and after interactions are done. To compute purity, each cluster is assigned to the class which is most frequent in it, and then the accuracy of this is measured. Users are assigned gold labels based on the most common label of all the nodes they are directly connected to in the graph. Results in Tab.~\ref{table:cluster_purity} show purity increases after interactions, showing interactions help learn better communities.

\subsection{Human Interaction Analysis/Examples}
\label{sec:human_annotation_analysis}
We analyze the interactions to determine what humans connected. We see humans make smart decisions in matching content preferences. Further, we show specific examples, demonstrating the ease, quickness, and lack of subjectivity of the interaction process. These details/ex. are in  App.~\ref{appendix:human_annotation_analysis}.

\subsection{Real World Use Case}
\label{sec:real_world_use_case}
As shown in Sec.~\ref{sec:experiments}, our interactive protocols enable rapidly (humans spent $\sim$3 min/sub-graph) learning better source factuality detection models for new events, even in the most challenging settings when there are no users, articles, or sources in common with prior data. This happens as contrary to providing additional labels, which can be time consuming and hard, interactions clear up content preferences, creating better social homophily and performance.

Specifically, in a real-world use case, interacting at training time learns a better model for the new event setting (Protocol 2 results on E2-1). In addition, this model would become even stronger as more interactions are performed, even without any further training, as seen in Protocol 3. Thus, when new news events happen, humans can interact on a few settings (our interaction sub-graphs) and our setup enables the model to amplify this knowledge to rapidly detect fake news sources on a large scale. 

%% file: summary.tex
We proposed an initial protocol to interactively build stronger information communities, applying it on source factuality detection. We focused on the early detection settings, where even strong  models can struggle. Our approach of finding sub-graphs and then interacting on them via 3 protocols enables minimal, quick human interactions to achieve significant performance improvements. We hypothesize that our interactive framework can generalize to other social media analysis tasks like bias or topic detection, and testing it is our future work. Additionally, we aim to scale up our interaction process, to include additional human interactions and types of interactions.

%% file: acknowledgements.tex
We thank the anonymous reviewers of this paper for all of their vital feedback. The project was funded by NSF CAREER award IIS-2048001 and IIS-2135573.